\documentclass[11pt]{article}

\usepackage[final]{acl}
\usepackage{times}
\usepackage{latexsym}
\usepackage[T1]{fontenc}
\usepackage[utf8]{inputenc}
\usepackage{microtype}
\usepackage{inconsolata}
\usepackage{graphicx}
\usepackage{booktabs}
\usepackage{amsmath}
\usepackage{amssymb}
\usepackage{xspace}
\usepackage{multirow}
\usepackage{enumitem}
\usepackage{float}
\usepackage{tabularx}

\newcommand{\sef}{\textsc{Sef}\xspace}

\title{From ``Thinking'' to ``Justifying'': Aligning High-Stakes Explainability with Professional Communication Standards}

\author{
  Chen Qian \\
  William \& Mary \\
  \texttt{cqian03@wm.edu} \\
  \And
  Yimeng Wang \\
  William \& Mary \\
  \texttt{ywang139@wm.edu} \\
  \And
  Yu Chen \\
  Anytime AI \\
  \texttt{ychen@anytime-ai.com} \\
  \AND
  Lingfei Wu \\
  Anytime AI \\
  \texttt{lwu@anytime-ai.com} \\
  \And
  Andreas Stathopoulos \\
  William \& Mary \\
  \texttt{axstat@wm.edu} \\
}

\begin{document}
\maketitle

\begin{abstract} 
Explainable AI (XAI) in high-stakes domains should help stakeholders trust and verify system outputs.
Yet Chain-of-Thought methods reason before concluding, and logical gaps or hallucinations can yield conclusions that do not reliably align with their rationale \citep{turpin2023language}.
Thus, we propose ``Result $\rightarrow$ Justify'', which constrains the \textit{output communication} to present a conclusion before its structured justification. We introduce \sef (Structured Explainability Framework), operationalizing professional conventions (e.g., CREAC, BLUF) via six metrics for structure and grounding. Experiments across four tasks in three domains validate this approach: all six metrics correlate with correctness ($r=0.20$--$0.42$; $p<0.001$), and \sef achieves 83.9\% accuracy (+5.3 over CoT). These results suggest structured justification can improve verifiability and may also improve reliability.
\end{abstract}

\section{Introduction}

As large language models (LLMs) are deployed in high-stakes domains, explainable AI (XAI) has become essential for helping humans assess model outputs \citep{bommasani2021opportunities}. 
Prior XAI work has progressed from post-hoc attribution \citep{ribeiro2016should, lundberg2017unified} to process-oriented approaches like Chain-of-Thought (CoT) \citep{wei2022chain, yao2023tree}. 
While CoT prompting elicits step-by-step reasoning before a conclusion, hallucinations or logical gaps can yield conclusions that misalign with their rationale \citep{turpin2023language, lanham2023measuring}.
Such misaligned or unfaithful explanations can hinder stakeholders' understanding of the model's reasoning.

To address this challenge, we draw on professional communication practices as a source of \textit{justification structure}, and use them to format explanations as verifiable justifications.
CREAC (Conclusion-Rule-Explanation-Analysis-Conclusion) in legal writing \citep{mangan2022creac} and BLUF (Bottom Line Up Front), a U.S. military communication standard widely adopted in business \citep{sehgal2016write}, are widely used templates for presenting a defensible conclusion and its supporting evidence. Legal theory frames this as the distinction between the \textit{context of discovery} (how one arrives at an answer) and the \textit{context of justification} (how one defends it) \citep{burton2007introduction}. We hypothesize that AI explanations in high-stakes domains may benefit from adopting this justification-oriented structure: ``\textbf{Result $\rightarrow$ Justify}.'' Crucially, \sef targets the \textit{context of justification} by constraining output communication rather than internal deliberation. 
Such perspective is supported by mathematical writing, where results precede polished proofs that may differ from the original derivation, emphasizing clarity and evaluability \citep{lamport2012write, aigner2010proofs}.

\begin{figure*}[t]
\centering
\includegraphics[width=\textwidth]{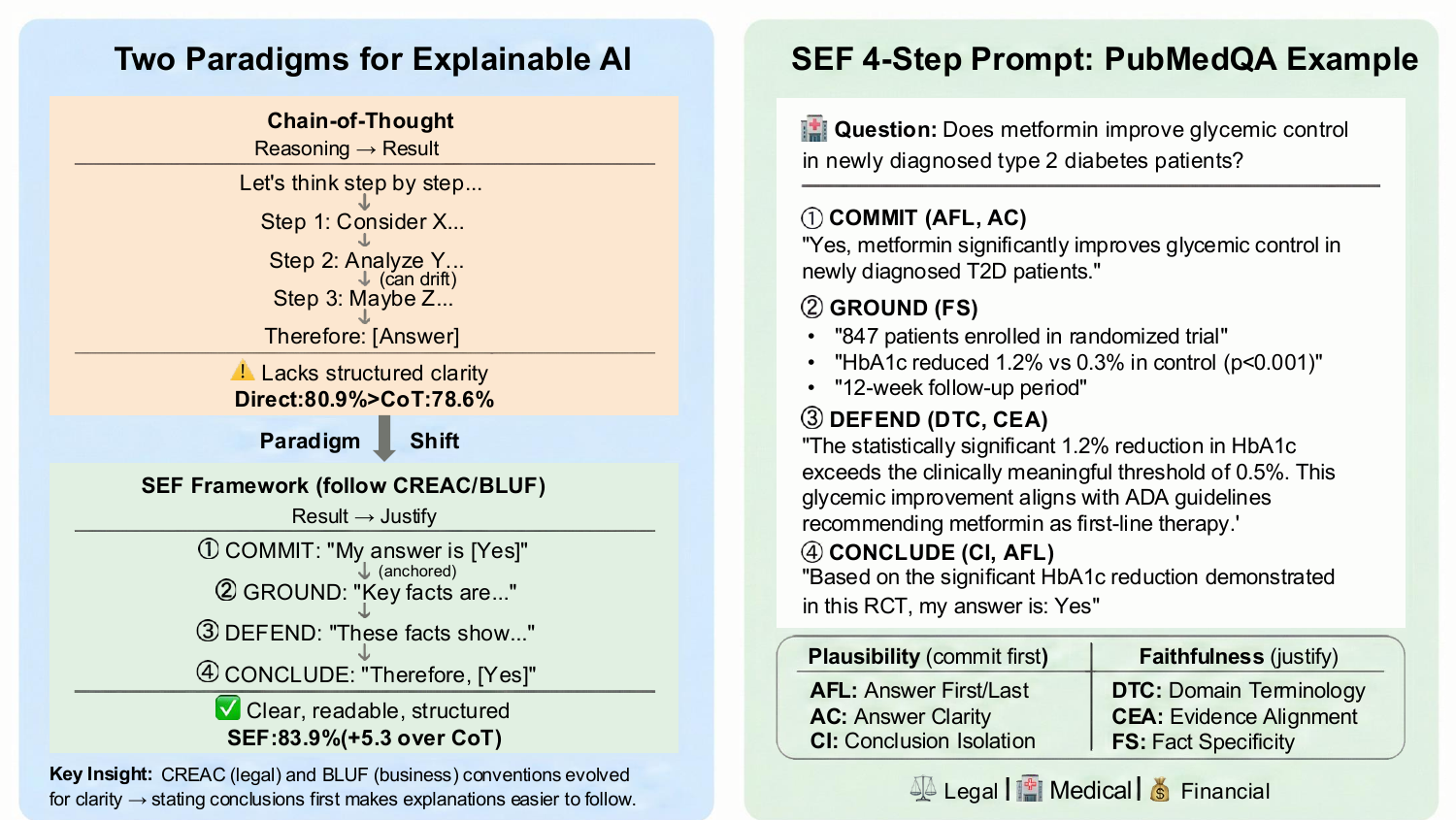}
\caption{\textbf{\sef: Structured Explainability via CREAC/BLUF Conventions.} \textbf{Left:} Paradigm comparison showing CoT (``Reasoning $\rightarrow$ Result'') vs.\ \sef (``Result $\rightarrow$ Justify''). CoT traces can be hard to verify (Direct 80.9\% $>$ CoT 78.6\%); \sef achieves 83.9\%. \textbf{Right:} PubMedQA example demonstrating the 4-step prompt (Commit, Ground, Defend, Conclude) with metric mappings.}
\label{fig:framework}
\vspace{-15pt}
\end{figure*}

We propose \sef (Structured Explainability Framework), which operationalizes CREAC and BLUF conventions through six metrics capturing \textit{plausibility} (structural clarity) and \textit{faithfulness} (evidence grounding) \citep{jacovi2020towards} (Figure~\ref{fig:framework}). While mechanistic interpretability examines internal activations, SEF acts as a behavioral probe: by measuring output structure, we infer how prompting constraints shape reasoning organization \citep{lanham2023measuring, lyu2023faithful}. Experiments across four tasks in three domains validate this approach:

\begin{enumerate}[leftmargin=*, noitemsep, topsep=0pt, partopsep=0pt]
    \item All six metrics correlate with correctness ($r=0.20$--$0.42$; $p<0.001$), suggesting that structured justifications align with more reliable outputs.
    \item \sef achieves 83.9\% accuracy (+5.3 over CoT). Notably, Direct (80.9\%) outperforms CoT (78.6\%), consistent with concerns about unconstrained reasoning.
    \item Ablations show the conclusion-first presentation scaffold is the main driver of accuracy: removing plausibility constraints causes the largest drop ($-$32.9), while removing faithfulness constraints reduces accuracy modestly ($-$4.3) but weakens domain grounding.
\end{enumerate}

\section{Related Work}

\paragraph{Process-Oriented Reasoning.} CoT \citep{wei2022chain} follows a ``Reasoning $\rightarrow$ Result'' paradigm but has known failures. \citet{turpin2023language} show CoT can be \textit{unfaithful}, offering plausible reasoning unrelated to actual computation. Similarly, \citet{lanham2023measuring} note models may reach correct answers via flawed steps. Remedies like self-consistency \citep{wang2022self}, Tree-of-Thought \citep{yao2023tree}, and Chain-of-Verification \citep{dhuliawala2024chain} improve accuracy but retain this paradigm without structural constraints, which reduces verifiability in high-stakes settings. We propose ``Result $\rightarrow$ Justify'' (like CREAC and BLUF): state the conclusion first, then build a structured, verifiable defense.

\paragraph{Explainability Evaluation.} Explanation evaluation separates faithfulness (actual model reasoning) from plausibility (human reasonableness) \citep{jacovi2020towards, wiegreffe2021teach}. Existing metrics (e.g., LIME, SHAP) focus on \textit{token-level} attribution \citep{ribeiro2016should, lundberg2017unified}. While these identify \textit{what} inputs influence predictions, they ignore \textit{how} the model structures arguments. Because current methods cannot distinguish coherent justification from a list of salient tokens, we propose metrics grounded in professional communication (e.g., evidence linking) to assess if AI explanations mirror expert practice.

\paragraph{Domain-Specific NLP.} Benchmarks exist for legal \citep{guha2023legalbench}, medical \citep{jin2019pubmedqa}, and financial \citep{malo2014good} NLP. We complement them with metrics that operationalize justification-oriented structure, which these domains value for review and accountability.

\section{\sef Framework}

\sef is dual-purpose: a prompting specification and a heuristic metric suite. Concretely, we use the metrics both to evaluate all methods and to design a constrained, sectioned output format for generation. Full heuristic definitions and the prompt template are in Appendix~\ref{app:impl} (App.~\ref{app:metrics_impl}--\ref{app:prompt_template}).

\paragraph{Six Justification Metrics.} We score each explanation with six heuristic metrics in $[0,1]$, grouped into two dimensions:

\noindent\textbf{Plausibility (commitment and sectioning).}
\begin{itemize}[leftmargin=*, noitemsep, topsep=0pt]
    \item \textit{Answer First/Last (AFL)}: answer appears in the first/last 200 characters (1.0 both, 0.5 one, 0.0 neither).
    \item \textit{Answer Clarity (AC)}: explicit answer statements via regex patterns (strong 1.0, medium 0.7, weak 0.3).
    \item \textit{Conclusion Isolation (CI)}: conclusion structurally separated (explicit conclusion headers 1.0; discourse markers 0.6).
\end{itemize}

\noindent\textbf{Faithfulness (justify with evidence).}
\begin{itemize}[leftmargin=*, noitemsep, topsep=0pt]
    \item \textit{Domain Terminology Consistency (DTC)}: counts curated domain terms (thresholded score 0.2--1.0).
    \item \textit{Conclusion--Evidence Alignment (CEA)}: detects evidence-linking language (e.g., ``based on'', ``according to'') plus analysis cues.
    \item \textit{Fact Specificity (FS)}: counts specificity indicators (numbers/quotes, enumerations) against vagueness markers.
\end{itemize}

\paragraph{Metric-informed prompting.} The metrics induce a 4-step schema: \textbf{Commit} (AFL, AC), \textbf{Ground} (FS), \textbf{Defend} (DTC, CEA), \textbf{Conclude} (CI, AFL). Unlike CoT, \sef commits first and constrains format to reduce drift.

\section{Experiments}
\label{sec:results}

\subsection{Setup}

\paragraph{Models.} We evaluate four open-weight 12--14B instruction-tuned models: DeepSeek-R1-Distill-Qwen-14B \citep{guo2025deepseek}, Gemma 3 12B \citep{team2024gemma}, Ministral-3-14B \citep{mistral2025mistral3}, and Qwen 2.5 14B \citep{bai2023qwen}. We use greedy decoding (temperature=0) via vLLM \citep{kwon2023efficient}.

\paragraph{Tasks.} We evaluate on four Yes/No classification tasks spanning legal \citep{guha2023legalbench}, medical \citep{jin2019pubmedqa}, and financial \citep{malo2014good} domains (1,618 test samples total; see Appendix~\ref{app:datasets} for details). The binary format makes accuracy a sufficient metric while enabling comparison of explanation quality through our six metrics.

\paragraph{Baselines.} We compare against six prompting strategies: (1) \textbf{Direct}, answer without explanation; (2) \textbf{Chain-of-Thought} \citep[CoT;][]{wei2022chain}, ``think step by step''; (3) \textbf{Tree-of-Thought} \citep[ToT;][]{yao2023tree}, explore multiple reasoning paths; (4) \textbf{Chain-of-Verification} \citep[CoVe;][]{dhuliawala2024chain}, verify reasoning steps; (5) \textbf{Vanilla RAG} \citep[V-RAG;][]{lewis2020retrieval}, retrieve relevant context; (6) \textbf{Self-RAG} \citep{asai2024self}, iterative retrieval with self-reflection. All baselines use the same greedy decoding and are evaluated on identical test sets; full prompt templates are in Appendix~\ref{app:baseline_prompts}.

\paragraph{Ablation Protocol.} For systematic ablation, we create eight \sef variants: six single-component ablations (removing AFL, AC, CI, DTC, CEA, or FS instructions), plus two dimension-level ablations (removing all Presentation or all Domain instructions). Each variant runs on all four models across all four tasks (64 experimental conditions total).

\subsection{Metric-Accuracy Correlation}

\begin{table}[t]
\centering
\small
\setlength{\tabcolsep}{4pt}
\begin{tabular}{llcc}
\toprule
\textbf{Dimension} & \textbf{Metric} & \textbf{Pearson $r$} & \textbf{$p$-value} \\
\midrule
Plausibility & AFL & 0.36 & $<.001$ \\
 & AC & 0.42 & $<.001$ \\
 & CI & 0.31 & $<.001$ \\
\midrule
Faithfulness & DTC & 0.20 & $<.001$ \\
 & CEA & 0.23 & $<.001$ \\
 & FS & 0.36 & $<.001$ \\
\bottomrule
\end{tabular}
\caption{Pearson correlation between each metric and prediction correctness ($n{=}90{,}608$ samples across all models, methods, and tasks). All metrics show significant correlations with accuracy ($p<0.001$). Plausibility metrics tend to show stronger effects ($r=0.31$--$0.42$) than faithfulness metrics ($r=0.20$--$0.36$).}
\label{tab:correlations}
\vspace{-12pt}
\end{table}

We first examine whether the metrics correlate with prediction accuracy. Crucially, we compute each metric on 90,608 outputs from \textit{all} methods (4 models $\times$ 7 methods $\times$ 4 tasks $\times$ variable samples). The correlations reflect general associations between structure and correctness, not artifacts of SEF-specific formatting.

Table~\ref{tab:correlations} shows Pearson correlations between each metric and prediction correctness. All six metrics correlate significantly with accuracy ($p<0.001$), supporting our hypothesis that CREAC/BLUF-style justification structure is associated with more reliable outputs. Both dimensions contribute: plausibility metrics ($r=0.31$--$0.42$) capture commitment and sectioning, while faithfulness metrics ($r=0.20$--$0.36$) capture domain grounding.

\subsection{Main Results}

\begin{table}[t]
\centering
\small
\setlength{\tabcolsep}{3pt}
\begin{tabular}{lccccc}
\toprule
\textbf{Method} & \textbf{FPB} & \textbf{CQA} & \textbf{Hear.} & \textbf{PMQ} & \textbf{Avg} \\
\midrule
Direct & 94.3 & 93.9 & 49.5 & 86.1 & 80.9 \\
CoT & 92.7 & 91.5 & 49.7 & 80.5 & 78.6 \\
ToT & 93.6 & 89.8 & 45.0 & 82.4 & 77.7 \\
CoVe & 88.9 & 90.2 & \underline{54.5} & 83.9 & 79.4 \\
V-RAG & 95.3 & 86.6 & 45.0 & 81.0 & 76.9 \\
Self-RAG & 76.5 & 88.0 & 48.1 & 76.9 & 72.4 \\
\midrule
\textbf{\sef} & \textbf{95.5} & \textbf{96.1} & \textbf{54.5} & \textbf{89.5} & \textbf{83.9} \\
\bottomrule
\end{tabular}
\caption{Accuracy (\%) averaged over four models. \sef achieves highest average accuracy (83.9\%), outperforming all baselines including CoT (+5.3 points). \textbf{Tasks:} FPB = Financial PhraseBank, CQA = ConsumerQA, Hear. = Hearsay, PMQ = PubMedQA.}
\label{tab:main}
\vspace{-5pt}
\end{table}

Table~\ref{tab:main} shows that \sef achieves 83.9\% average accuracy, outperforming CoT by 5.3 points (+9.0 on PubMedQA, +4.6 on ConsumerQA, +2.8 on FPB). On Hearsay, \sef matches the best baseline (CoVe) at 54.5\%.

Notably, Direct prompting (80.9\%) outperforms CoT (78.6\%) and most reasoning methods, a counter-intuitive finding consistent with recent work showing that CoT's free-form reasoning traces can introduce errors through drift or unfaithful explanations \citep{turpin2023language}. \sef outperforms even Direct (+3.0 points), suggesting that \textit{structured} justification adds value beyond mere answer commitment.

\subsection{Ablation Study}

\begin{table}[t]
\centering
\small
\setlength{\tabcolsep}{3pt}
\begin{tabular}{lcccccc}
\toprule
\textbf{Variant} & \textbf{FPB} & \textbf{CQA} & \textbf{Hear.} & \textbf{PMQ} & \textbf{Avg} & \textbf{$\Delta$} \\
\midrule
\sef (Full) & 95.5 & 96.1 & 54.5 & 89.5 & 83.9 & -- \\
\midrule
w/o AFL & 94.5 & 92.3 & 41.3 & 86.9 & 78.8 & $-$5.1 \\
w/o AC & 90.5 & 91.0 & 47.8 & 85.0 & 78.6 & $-$5.3 \\
w/o CI & 70.9 & 68.9 & 49.3 & 56.7 & 61.4 & $-$22.5 \\
\midrule
w/o DTC & 94.0 & 91.4 & 48.3 & 84.8 & 79.6 & $-$4.3 \\
w/o CEA & 90.5 & 90.5 & 54.7 & 84.5 & 80.0 & $-$3.9 \\
w/o FS & 90.8 & 91.6 & 50.4 & 84.8 & 79.4 & $-$4.5 \\
\midrule
w/o Pres. & 35.1 & 67.5 & 53.8 & 47.8 & 51.0 & $-$32.9 \\
w/o Domain & 92.3 & 90.5 & 50.9 & 84.5 & 79.6 & $-$4.3 \\
\bottomrule
\end{tabular}
\caption{Ablation results: accuracy (\%) when removing individual components or entire dimensions. Plausibility components (especially CI) tend to cause larger drops than faithfulness components, consistent with correlation patterns in Table~\ref{tab:correlations}. \textbf{Notation:} w/o = without; Pres. = Plausibility (AFL+AC+CI); Domain = Faithfulness (DTC+CEA+FS).}
\label{tab:ablation}
\vspace{-12pt}
\end{table}

Table~\ref{tab:ablation} reveals how constraints contribute to verifiable justifications:

\paragraph{Plausibility enables structured presentation.} Without plausibility constraints, SEF degrades to unstructured reasoning, causing a massive accuracy drop ($-$32.9). This underscores that the explicit structural constraints are the primary driver of performance.

\paragraph{Faithfulness enables domain grounding.} Removing faithfulness constraints reduces accuracy modestly ($-$4.3) but results in generic explanations. The impact is task-dependent: DTC is vital for Hearsay ($-$6.2) due to legal precision, while FS drives performance on PubMedQA by enforcing clinical evidence citation.

\paragraph{Both dimensions are necessary.} Plausibility provides the structural scaffold; faithfulness fills it with domain-appropriate content. Neither alone yields fully verifiable justifications.

\section{Discussion}

\paragraph{Accuracy as behavioral validation.} Direct (80.9\%) outperforming CoT (78.6\%) is consistent with work questioning CoT faithfulness \citep{turpin2023language}, suggesting that CoT's free-form reasoning traces may sometimes hurt rather than help. \sef (83.9\%) combines answer commitment (like Direct) with structured justification (unlike Direct's lack of explanation, and unlike CoT's unconstrained traces). This accuracy pattern validates that CREAC/BLUF-style structure \citep{mangan2022creac, sehgal2016write} reflects effective output organization, not just convention.

\paragraph{What output structure reveals.} The systematic correlation between structure and accuracy ($r=0.20$--$0.42$) suggests prompting constraints shape how models organize their outputs. Imposing conclusion-first structure is associated with higher accuracy on average. This behavioral analysis complements mechanistic interpretability by highlighting \textit{which output organization patterns} coincide with reliable outputs.

\paragraph{Implications.} For high-stakes domains, explanations may benefit from mirroring professional practice: structured defenses of committed conclusions, rather than reasoning traces hoping to find answers. Our results suggest this improves both interpretability and reliability.

\section{Conclusion}

We introduced \sef, which structures AI explanations via CREAC and BLUF conventions. This ``Result $\rightarrow$ Justify'' paradigm states the conclusion upfront, then builds a structured defense as a \emph{presentation} scaffold. \sef achieves 83.9\% accuracy (+5.3 over CoT), suggesting verifiable justifications may also improve reliability. Additionally, our six metrics provide scalable signals of quality to enable verification-oriented evaluation. For high-stakes domains, conclusion-first justification supports more trustworthy model use.

\section*{Limitations}

\paragraph{Proxy metrics.} Our six metrics are scalable heuristics that approximate professional communication conventions, not deep domain soundness. They measure \textit{structural} compliance (e.g., presence of conclusion headers, domain terms) rather than \textit{semantic} correctness of the justification itself. A response can score well while containing factual errors. We mitigate this concern by computing correlations across \textit{all} methods (not just SEF), showing that structure correlates with accuracy generally. Importantly, even when the answer is incorrect, structured justifications improve readability: the human-friendly format makes it easier for reviewers to identify errors in the reasoning. These automatic metrics serve as a scalable first-pass filter; human expert evaluation to assess semantic faithfulness remains future work.

\paragraph{Statistical interpretation.} Our per-sample correlations quantify association (effect size $r$), not causality, and samples may not be independent due to shared prompts, models, and datasets. We therefore interpret them descriptively and complement them with ablations that intervene on prompting constraints.

\paragraph{Scope and artifacts.} We validate on Yes/No tasks across three domains with 12--14B instruction-tuned models. Extending to multi-class settings, open-ended generation, additional domains, or larger models remains future work. We use public datasets and open-weight models under their stated licenses/terms and do not redistribute third-party datasets or model weights.

\section*{Ethical Considerations}

\paragraph{Persuasive hallucinations.} Structured justifications may make incorrect answers more convincing. Any use of \sef should include human oversight and avoid presenting generated justifications as professional advice or as evidence of correctness.

\paragraph{Professional boundaries.} \sef should not replace professional judgment in legal, medical, or financial decisions. Our 83.9\% accuracy implies substantial error rates.

\paragraph{Privacy and data use.} We do not collect new personal data or deploy systems. We use publicly released research benchmarks; users should respect dataset terms and avoid applying the framework to sensitive or identifiable records without appropriate approvals and safeguards.

\paragraph{Environmental impact.} We do not train new models. Experiments use existing 12--14B models with deterministic decoding, but compute remains non-trivial; releasing code and configurations is intended to reduce redundant reruns.

\bibliography{main}

\clearpage
\appendix

\section{Implementation Details}
\label{app:impl}

\paragraph{Reproducibility note.} \sef's metrics are \textbf{rule-based heuristics} (regex/pattern matching), not model-based judges. This makes the evaluation lightweight and deterministic. These metrics assess structural compliance rather than semantic correctness; their value lies in producing human-readable outputs that facilitate expert review. We summarize the exact scoring rules below.

\subsection{Metric Computation (Regex/Heuristics)}
\label{app:metrics_impl}

\paragraph{Inputs.} Each metric consumes an explanation string $e$ and a predicted binary answer $a \in \{\texttt{Yes},\texttt{No}\}$. All matching is performed on a lowercased copy of $e$.

\paragraph{AFL (Answer First/Last).} Let $e_{head}$ be the first 200 characters of $e$ and $e_{tail}$ the last 200. AFL is 1.0 if $a$ occurs in both $e_{head}$ and $e_{tail}$, 0.5 if it occurs in exactly one, else 0.0. Occurrence is checked by direct substring match or a word-boundary match for \texttt{Yes}/\texttt{No}.

\paragraph{AC (Answer Clarity).} We detect explicit answer statements with regexes. If any \emph{strong} pattern matches, AC=1.0; else if any \emph{medium} pattern matches, AC=0.7; else if $a$ appears anywhere in $e$, AC=0.3; else 0.0. Strong patterns capture explicit declarations (e.g., ``My answer is: Yes'', ``Final answer: No'', ``The answer is: Yes'', or ``Answer: Yes''); medium patterns capture discourse markers that directly precede a Yes/No conclusion (e.g., ``therefore, Yes'' or ``in conclusion, No'').

\paragraph{CI (Conclusion Isolation).} CI=1.0 if the output contains an explicit conclusion header (e.g., ``CONCLUSION:'' / ``Conclusion:''); CI=0.6 if it contains common concluding markers (e.g., ``in conclusion'', ``to conclude'', ``in summary'', ``therefore'', ``thus''); else 0.0.

\paragraph{DTC (Domain Terminology Consistency).} We count occurrences of curated, domain-specific terms (legal/medical/financial lexicons). If no lexicon is defined, DTC=0.5. Otherwise, if the term-count is $\ge 5$, DTC=1.0; if $\ge 3$, DTC=0.8; if $\ge 1$, DTC=0.5; else DTC=0.2.

\paragraph{CEA (Conclusion--Evidence Alignment).} We count evidence-linking cues and analysis cues. If link-cue count $\ge 3$ and analysis-cue count $\ge 1$, CEA=1.0; else if link-cue count $\ge 2$, CEA=0.8; else if link-cue count $\ge 1$, CEA=0.5; else if analysis-cue count $\ge 1$, CEA=0.3; else 0.0. Cues include phrases such as ``based on'', ``according to'', ``this suggests'', and explicit ``ANALYSIS'' sectioning.

\paragraph{FS (Fact Specificity).} We count specificity indicators and vagueness indicators. FS=1.0 if specificity count $\ge 4$ and vagueness count $\le 1$; else if specificity count $\ge 3$, FS=0.8; else if $\ge 2$, FS=0.6; else if $\ge 1$, FS=0.4; else 0.2. Specificity cues include numbers and quotes, enumerations (e.g., ``first/second''), and explicit fact headers (e.g., ``KEY FACTS''); vagueness cues include hedges (e.g., ``generally'', ``possibly'').

\begin{table}[t]
\centering
\small
\setlength{\tabcolsep}{5pt}
\begin{tabularx}{\columnwidth}{lX}
\toprule
\textbf{Metric} & \textbf{Heuristic scoring rule (summary)} \\
\midrule
\textbf{AFL} & Answer appears in both the first and last 200 characters (1.0); only one side (0.5); neither (0.0). \\
\textbf{AC} & Explicit answer declaration (1.0); weaker conclusion marker preceding Yes/No (0.7); answer appears anywhere (0.3); otherwise (0.0). \\
\textbf{CI} & Explicit conclusion header present (1.0); concluding discourse markers present (0.6); otherwise (0.0). \\
\textbf{DTC} & Domain term count $\ge 5$ (1.0); $\ge 3$ (0.8); $\ge 1$ (0.5); otherwise (0.2). \\
\textbf{CEA} & Link-cue count $\ge 3$ and analysis-cue count $\ge 1$ (1.0); link cues $\ge 2$ (0.8); $\ge 1$ (0.5); analysis cues $\ge 1$ (0.3); otherwise (0.0). \\
\textbf{FS} & Specificity count $\ge 4$ and vagueness count $\le 1$ (1.0); specificity $\ge 3$ (0.8); $\ge 2$ (0.6); $\ge 1$ (0.4); otherwise (0.2). \\
\bottomrule
\end{tabularx}
\caption{Summary of the six SEF metrics and their heuristic scoring rules.}
\label{tab:metric_summary}
\end{table}

\subsection{SEF Prompt Template}
\label{app:prompt_template}

\paragraph{Overview.} Our SEF prompting method enforces a conclusion-first structured format with four labeled sections. Table~\ref{tab:sef_prompt_template} shows the prompt template (domain role/instructions + sectioned format) used in our experiments; ablations remove the corresponding instructions/sections.

\begin{table*}[t]
\centering
\small
\setlength{\tabcolsep}{6pt}
\begin{tabularx}{\textwidth}{p{0.22\textwidth}Xp{0.22\textwidth}}
\toprule
\textbf{Prompt component} & \textbf{Template text (verbatim headers)} & \textbf{Metrics targeted} \\
\midrule
Role instruction & \texttt{[Domain role] Analyze the following question with precision and clarity.} & -- \\
\addlinespace[2pt]
Domain terminology (optional) & \texttt{Use precise [legal/medical/financial] terminology consistently.} & DTC \\
\addlinespace[2pt]
Fact specificity (optional) & \texttt{Cite specific facts from the context, not vague generalizations.} & FS \\
\addlinespace[2pt]
Context + question & \texttt{Context: ...} \newline \texttt{Question: ...} & -- \\
\midrule
Section 1: Commit & \texttt{ANSWER PREVIEW: State your answer upfront (Yes or No).} & AFL (first), AC \\
\addlinespace[2pt]
Section 2: Ground & \texttt{KEY FACTS: List 2-3 specific facts from the context that are most relevant.} & FS \\
\addlinespace[2pt]
Section 3: Defend & \texttt{ANALYSIS:} \newline \texttt{- Use precise domain terminology} \newline \texttt{- Explain how each fact supports your answer} & DTC, CEA \\
\addlinespace[2pt]
Section 4: Conclude & \texttt{CONCLUSION:} \newline \texttt{- State your final answer clearly and unambiguously} \newline \texttt{- Summarize the key evidence supporting your answer} \newline \texttt{- End with: My answer is: [Yes/No]} & CI, AFL (last), AC \\
\bottomrule
\end{tabularx}
\caption{SEF prompt template used in our experiments.}
\label{tab:sef_prompt_template}
\end{table*}

\subsection{Baseline Prompt Templates}
\label{app:baseline_prompts}

For reproducibility, we document the core prompt structures for each baseline. All methods receive the same \texttt{Context} and \texttt{Question} inputs; only the instruction wrapper differs.

\paragraph{Direct.} Minimal prompt requesting only the answer:
\begin{quote}
\small\ttfamily
Answer the following question.\\
Context: [context]\\
Question: [question]\\
Provide only the answer (Yes/No).\\
Answer:
\end{quote}

\paragraph{Chain-of-Thought (CoT).} Free-form reasoning triggered by ``think step by step'':
\begin{quote}
\small\ttfamily
Given the following context and question, let's think step by step.\\
Context: [context]\\
Question: [question]\\
Let's think step by step:\\
1. First, let me understand the context and relevant rules.\\
2. Then, I'll analyze how they apply to this question.\\
3. Finally, I'll determine the correct answer.\\
Step-by-step reasoning:
\end{quote}

\paragraph{Tree-of-Thought (ToT).} Three-stage process: (1) generate $k{=}3$ initial approaches, (2) develop each into full analysis, (3) evaluate and select best path.
\begin{quote}
\small\ttfamily
\textbf{Stage 1:} Propose 3 different analytical approaches (1-2 sentences each).\\
\textbf{Stage 2:} For each approach: ``Develop this into a complete analysis... then state your final answer.''\\
\textbf{Stage 3:} ``Which path (1, 2, or 3) provides the most thorough and accurate analysis?''
\end{quote}

\paragraph{Chain-of-Verification (CoVe).} Four-stage process with self-verification:
\begin{quote}
\small\ttfamily
\textbf{Stage 1:} ``Answer the question with reasoning.''\\
\textbf{Stage 2:} ``Generate 3 verification questions to check accuracy.''\\
\textbf{Stage 3:} Answer each verification question against the context.\\
\textbf{Stage 4:} ``Based on initial analysis and verification, provide your final answer (incorporate any corrections).''
\end{quote}

\paragraph{Vanilla RAG (V-RAG).} Retrieve-then-generate with top-$k{=}3$ passages:
\begin{quote}
\small\ttfamily
Answer the question using the provided context.\\
Retrieved Context: [top-k passages by keyword overlap]\\
Question: [question]\\
Based on the retrieved context, provide your reasoning and answer.
\end{quote}

\paragraph{Self-RAG.} Multi-stage retrieval with reflection and self-critique:
\begin{quote}
\small\ttfamily
\textbf{Stage 1:} Assess if retrieval is needed (RETRIEVE or USE\_FULL).\\
\textbf{Stage 2:} If retrieving, filter passages by relevance (RELEVANT or NOT\_RELEVANT).\\
\textbf{Stage 3:} Generate with reflection: ``[Reflection: ...] [Generation: ...] [Answer: ...]''\\
\textbf{Stage 4:} Self-critique for accuracy (NEEDS\_REFINEMENT or SUFFICIENT).\\
\textbf{Stage 5:} If needed, refine based on critique.
\end{quote}

\subsection{Datasets and Input Formatting}
\label{app:datasets}

\paragraph{Tasks and sources.} We evaluate on four binary classification tasks spanning three high-stakes domains, totaling 1,618 test samples:
\begin{itemize}[leftmargin=*, noitemsep, topsep=0pt]
    \item \textbf{Legal} \citep{guha2023legalbench}: ConsumerQA (396 samples), identifying unfair terms in consumer contracts; Hearsay (94 samples), applying Federal Rule of Evidence 801(c).
    \item \textbf{Medical}: PubMedQA \citep{jin2019pubmedqa} (1,000 samples), answering biomedical research questions based on abstracts.
    \item \textbf{Financial}: FPB \citep{malo2014good} (128 samples), classifying sentiment in financial news phrases.
\end{itemize}

\paragraph{Evaluation splits.} We use the evaluation splits provided by each benchmark.

\paragraph{Input formatting.} Each example is formatted into a \texttt{Context} field (the provided passage/document) and a \texttt{Question} field. All prompting methods receive the same inputs; only the instruction/prompt wrapper differs across methods.

\subsection{Exact Model Identifiers}
\label{app:model_ids}
\noindent We report paper-friendly names in the main text; the exact vLLM/HF identifiers used by our experiment runner are:
\begin{itemize}[leftmargin=*, noitemsep, topsep=0pt]
    \item DeepSeek-R1-Distill-Qwen-14B: \texttt{deepseek-ai/DeepSeek-R1-Distill-Qwen-14B}
    \item Gemma 3 12B: \texttt{google/gemma-3-12b-it}
    \item Ministral-3-14B: \texttt{mistralai/Ministral-3-14B-Instruct-2512}
    \item Qwen 2.5 14B Instruct: \texttt{Qwen/Qwen2.5-14B-Instruct}
\end{itemize}

\end{document}